\documentclass[letterpaper]{article} % DO NOT CHANGE THIS
\usepackage{aaai25}  % DO NOT CHANGE THIS
\usepackage{times}  % DO NOT CHANGE THIS
\usepackage{helvet}  % DO NOT CHANGE THIS
\usepackage{courier}  % DO NOT CHANGE THIS
\usepackage[hyphens]{url}  % DO NOT CHANGE THIS
\usepackage{graphicx} % DO NOT CHANGE THIS
\usepackage{natbib}  % DO NOT CHANGE THIS AND DO NOT ADD ANY OPTIONS TO IT
\usepackage{caption} % DO NOT CHANGE THIS AND DO NOT ADD ANY OPTIONS TO IT
\usepackage{algorithm}
\usepackage{algorithmic}

\usepackage{amsmath}
\usepackage{amssymb}
\usepackage{mathtools}
\usepackage{amsthm}

\usepackage[capitalize,noabbrev]{cleveref}
\usepackage[table,xcdraw]{xcolor}

\usepackage{newfloat}
\usepackage{listings}
\DeclareCaptionStyle{ruled}{labelfont=normalfont,labelsep=colon,strut=off} % DO NOT CHANGE THIS
\floatstyle{ruled}
\newfloat{listing}{tb}{lst}{}
\floatname{listing}{Listing}

\title{CTD4 - a Deep Continuous Distributional Actor-Critic Agent\\ with a Kalman Fusion of Multiple Critics}
\author {
    David Valencia\textsuperscript{\rm 1},
    Henry Williams\textsuperscript{\rm 1},
    Yuning Xing\textsuperscript{\rm 1},\\
    Trevor Gee\textsuperscript{\rm 1},
    Bruce A MacDonald\textsuperscript{\rm 1},
    Minas Liarokapis \textsuperscript{\rm 2}
}
\affiliations {
    \textsuperscript{\rm 1}Centre for Automation and Robotic Engineering Science, The University of Auckland, Auckland, New Zealand\\
    \textsuperscript{\rm 2}New Dexterity Lab, Auckland, New Zealand\\
    \{dval035, yxin683\}@aucklanduni.ac.nz, \{henry.williams, t.gee, b.macdonald, minas.liarokapis\}@auckland.ac.nz
}

\begin{document}

\maketitle

\begin{abstract}
    Categorical Distributional Reinforcement Learning (CDRL) has demonstrated superior sample efficiency in learning complex tasks compared to conventional Reinforcement Learning (RL) approaches. However, the practical application of CDRL is encumbered by challenging projection steps, detailed parameter tuning, and domain knowledge. This paper addresses these challenges by introducing a pioneering Continuous Distributional Model-Free RL algorithm tailored for continuous action spaces. The proposed algorithm simplifies the implementation of distributional RL, adopting an actor-critic architecture wherein the critic outputs a continuous probability distribution. Additionally, we propose an ensemble of multiple critics fused through a Kalman fusion mechanism to mitigate overestimation bias. Through a series of experiments, we validate that our proposed method provides a sample-efficient solution for executing complex continuous-control tasks.
\end{abstract}

\section{Introduction}
    % Talk about RL and distributions:
        The real world has a stochastic nature characterized by different levels of uncertainty that are complex to model or predict with deterministic systems. Thus, the adoption of stochastic strategies becomes imperative to obtain accurate information on environmental dynamics. This necessity is particularly important in RL,  where an intelligent agent is tasked with learning effective behaviours in the presence of noisy, stochastic environments. In addressing this challenge, Distributional RL emerges as a viable solution, offering a framework to navigate the stochastic nature of the world effectively, an aspect that traditional RL methods struggle to handle directly.

        Currently, most of the state-of-the-art RL algorithms use the expectations of returns $Q^{\pi}$ (as a scalar value) for each action-state pair following a policy $\pi$ \cite{fujimoto2018addressing, schulman2017proximal, haarnoja2018soft}. RL requires an accurate approximation of the Q-function to guarantee sample efficiency and stability. An adequate approximation is essential to calculate the Temporal Difference (TD) error or action selection, whether in policy-based or value-based approaches \cite{kuznetsov2020controlling}. Therefore, better and more stable approximations of the Q-function are needed.

        Distributional RL presents the capability to learn a better approximation of the Q-function as a complete distribution as well as the intrinsic behaviours and randomness associated with the environment and the policy. By learning a distribution of the returns rather than its expectations, policies can be learned more efficiently and achieve higher performance \cite{bellemare2017distributional, choi2019distributional}.

    % What are the problems: 

        % 1) Categorical distributions are complex to implement 
        The works of \cite{lyle2019comparative}, \cite{rowland2018analysis}, \cite{bellemare2017distributional, choi2019distributional} are examples of related literature that use distributions in RL for discrete control problems, where the output of the Q-function approximator is a categorical distribution. This type of distribution is usually highly dependent on several hyperparameters, such as the number of categories or bounding values. Furthermore, to get acceptable performance and reduce the approximation error, task-specific knowledge with additional steps such as projections, truncations, or auxiliary predictions are needed \cite{bellemare2017distributional}. Although the results may outperform traditional deterministic RL approaches, computing and training on categorical distributions are often complex.
        
        % 2) overestimation
        Likewise, either in traditional RL or distributional RL, the overestimation bias is one of the primary obstacles to accurate policy learning. When estimating a Q-function, an overestimation is inevitable due to the imperfect nature of the estimator. That is, the approximation may be significantly higher than the actual value. Prior studies addressed this problem through an ensemble of various Q-approximators in multiple ways. 

        For example, \cite{fujimoto2018addressing} and \cite{lan2020maxmin} address overestimation in control problems by utilizing minimum values from multiple network approximations. Similarly, \cite{kuznetsov2020controlling} and \cite{chen2021randomized} enhance Q estimation stability through ensemble learning and averaging techniques.
        
        We argue that taking the minimum value among the $Q$ approximations is not the best option since the opposite effect of overestimation could happen, i.e. underestimation, ending up in discouragement of exploration \cite{lan2020maxmin}. Additionally, using the minimum value among the approximations makes the ensemble lose its significance since the other values are ignored. Taking the average among the approximations is not optimal either, especially when working with distributions of returns. Weighting all of the distributions the same when they are not equally good could result in sub-optimal or incorrect approximations.
        
        % How we solve the problems 
        %----------
        In this work, to mitigate the issues of CDRL and overestimation bias, we present a novel Continuous Distributional Reinforcement Learning algorithm called Continuous Twin Delayed Distributed Deep Deterministic Policy Gradient \textbf{CTD4}. We extend and transform TD3, from a deterministic actor-critic method to a continuous distributional actor-critic approach. This transformation preserves the simplicity of the original implementation while incorporating the advantages of ensemble distributions to enhance the learning of an RL control policy for  \textbf{continuous action problems}.
        
        We propose using a normal distribution to parameterize the approximation of the return distribution.  With the normal distribution described by its mean $\mu$ and standard deviation $\sigma$, the disjoint support issues of categorical distributions and the high dependency on task-specific hyperparameters can be avoided. In the same way, no customized loss/distance metrics are needed. Furthermore, to alleviate the overestimation, we propose an ensemble of the distributional Q-approximators fused through a Kalman method, preserving the power of the ensemble without the need for complex tuning parameters. The paper contributions are: 
        
    %------------------
    % Paper contributions
    %-----------
        \begin{enumerate}
            \item Introduce a novel continuous distributional Actor-Critic RL framework tailored for continuous action domains. This framework streamlines prior distributional RL methods by eliminating the reliance on categorical representations less suited for continuous spaces while concurrently improving sample efficiency.
            
            \item Mitigation of overestimation bias through the utilization of an ensemble of continuous distributed critics.
            
            \item Fuse the ensemble of distributional critics efficiently through a Kalman fusion approach, maximizing the collective impact of the ensemble.
            
        \end{enumerate}
        %-----------

\section{Related Work}

    Learning a complete distribution of returns for each action $a$ and state $s$ pair instead of its mean $Q^{\pi}(s, a)$ has been studied and proven to be an effective solution that significantly improves the performance of RL agents \cite{bellemare2023distributional, lyle2019comparative, rowland2018analysis, bellemare2017distributional}. After the introduction of C51 by \cite{bellemare2017distributional}, considered the first distributional RL algorithm, several studies have been done, and research on distributional RL emerged. 
    
    %C51
    C51 proposes to learn a categorical probability distribution of the returns for each action (for a discrete action space), where the distribution range and the number of finite categories are empirically selected.  A concern with C51  is minimizing the distance (viewed as a cross-entropy loss) between the categorical predicted distribution and the categorical target distribution, but this is impractical due to the disjoint of the categories. The authors solve this by projecting the Bellman update, which aligns the distributions.
    
    % QR-DQN, % IQN
    Using C51 as a baseline, QR-DQN uses quantile regression to automatically transpose and adjust the distributions' locations \cite{dabney2018distributional}. This allows the minimisation of the Wasserstein distance to the target distribution. QR-DQN is extended by including an Implicit Quantile Network (IQN) in \cite{dabney2018implicit}. The authors propose to learn the full quantile values through the network. Therefore, the return distribution is controlled by the size of the IQN. In this work, the output is not a distribution as C51 or QR-DQN; it re-parameterises samples from a base distribution to the respective quantile values of a target distribution.
    
    % MoG-DQN
    To avoid the categorical distribution's disjoint problem \cite{choi2019distributional} propose MoG-DQN an RL method using a mixture of Gaussians to parameterize the value distribution. The authors also present a customised distance metric since the cross-entropy between mixtures could be analytically impractical. MoG-DQN solves the distribution's disjoint problem. However, employing a mixture of Gaussians could be computationally intense, where selecting the right number of Gaussians is an additional hand-tuned parameter where the parameterisation is often numerically unstable. 

    The previously mentioned works can operate \textbf{in discrete action spaces only} and were tested predominantly on Atari games. Even when the results presented show a substantial improvement (against DQN \cite{mnih2013playing} mostly), the acceptable performance is restricted to task/domain-specific configurations, hyperparameters auxiliary projection functions or extra networks. 
    
    %____________________________%
    % D4PG
    Distributional RL has also been presented for continuous control problems. D4PG \cite{barth2018distributed} where the authors use a categorical distributional critic based on C51, along with a prioritizing experience replay buffer and an N-step reward discount sum.
    % TQC
    Another categorical distributed  RL method capable of working with continuous control problems is TQC, presented by \cite{kuznetsov2020controlling}. The authors present an actor-multiple-critics method using C51 and QR-DQN as a baseline. An ensemble of critics is used to alleviate the overestimation in the predicted distributions. Each critic produces a truncated category distribution, followed by dropping several atoms. The average between the ensemble is taken to generate the final approximation distribution.
    % DSAC
    A comparable approach using distributions is presented in \cite{duan2021distributional}. The authors introduce an extension of SAC called DSAC, along with TD4, a variation of TD3. Both approaches utilize continuous distributions to mitigate overestimation biases. DSAC models the Q-values as distributions using Soft Policy Improvement. However, the paper still derives a single Q-value estimate, which remains the primary source of overestimation. Additionally, the authors employ the PABAL architecture, which parallelized training by incorporating four learners, which is computationally demanding.
    %____________________________%
    
    % Summary
    Despite the promising results demonstrated in these works, recent literature on distributional RL lacks alternatives tailored for continuous control problems. Previous proposals are computationally expensive with complex and slow training steps and/or are still based on categorical distributions \cite{kuznetsov2020controlling}. 
    %____________________________%
    %____________________________%
    CDRL encompasses issues such as projection steps, customized loss/metric functions, or alternative techniques characterized by a significant reliance on tuning parameters. We tackle the majority of these issues, presenting a novel, easy-to-train, scalable, and stable method that is able to solve control tasks in continuous action space. As far as we are aware, we present the first distributional RL algorithm that uses continuous distributions in an actor-critics architecture for continuous control tasks addressing overestimation through a fusion of an ensemble of multiple distributed critics using a Kalman approach. 
   %____________________________%

\section{Background}

    \begin{figure}
        \centering
            \includegraphics[width=\linewidth]{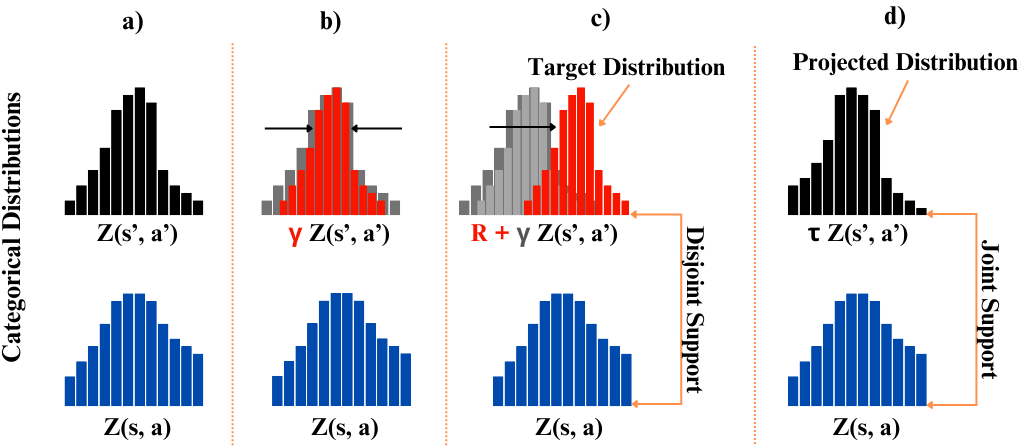}
            \caption{Categorical Distribution Computing Problem. The top part represents the sequence of the target distribution being shrunk and shifted by the discount factor $\gamma$ and Reward $R$, respectively. Then, it is projected to the original support through probabilities inversely proportional to the distance from the nearest support.  The bottom part describes the current distribution used to calculate the distance error.}
            \label{fig:categorical}
    \end{figure}

    %\subsection{Distributional RL}
    \begin{figure}
        \centering
            \includegraphics[width=\linewidth]{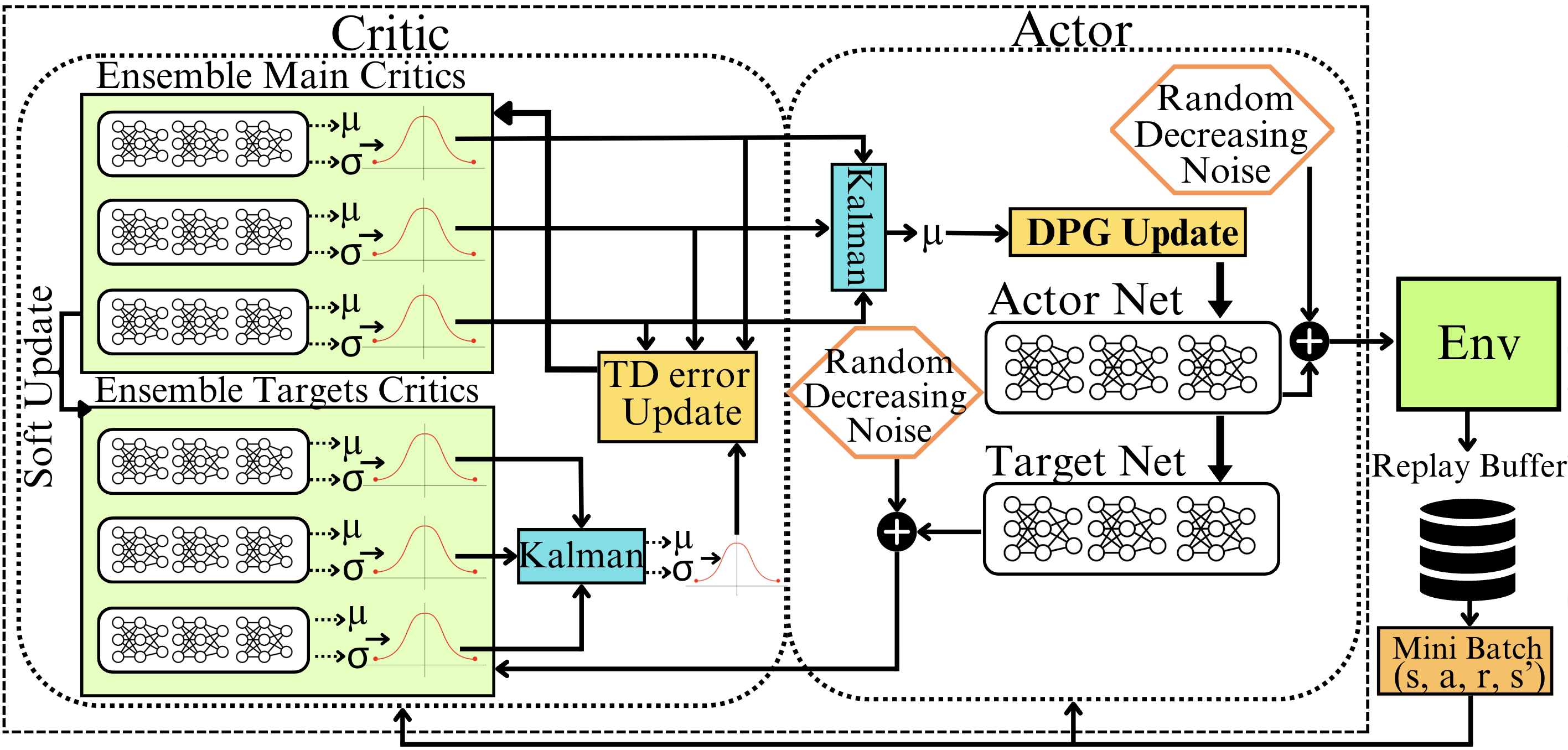}
            \caption{ 
            Diagram of the proposed continuous distributional method. The CTD4 architecture network consists of an actor network and $N$ critic networks. Each critic network comprising the ensemble consists of two hidden fully connected layers with 256 nodes, each with ReLU as an activation function. A linear layer is added for each output layer, i.e., the mean $\mu$ and standard deviation $\sigma$.  A softplus activation function is also included for $\sigma$ to guarantee positive values. The actor has Tanh as an activation function for the output layer along with two hidden fully connected layers with 256 nodes, each with ReLU.}
            \label{fig:diagram}
    \end{figure}
    
    % concepts Distributional RL but more technical/ math
    In standard RL, the Q-function is described as a scalar value, i.e. the expected return $ Q^\pi(s, a) \coloneqq E [\sum_{t=0}^{\infty} \gamma^t R(s_t, a_t)] $. However, the expectation is removed from the return equation in distributional RL. That is, distributional RL focuses on learning the full distribution of the random variable $Z^\pi (s, a)  \coloneqq  \sum_{t=0}^{\infty} \gamma^t R(s_t, a_t) $. Therefore, the distributed $Z^\pi$ function takes the central role and replaces the scalar $ Q^\pi$  function. The Bellman equation for distributional RL is presented as $Z^\pi (s, a) \stackrel{D}{=} R(s, a) + \gamma Z^\pi (s', a')$. 
    What primarily differentiates distributional RL from standard RL is the loss function. As with the scalar setting, the TD error can be estimated as the difference between the estimation and target values. However, in the case of distributional RL, the TD error is calculated as the distance between the current estimation distribution and the target distribution, presented as the cross-entropy term of the Kullback-Leibler (KL) divergence $\delta = D_{KL}[Z^\pi (s, a) - Y(s', a')]$ where the target distribution is defined as $Y(s', a') = R + \gamma Z^\pi (s', a')$.  

    In CDRL, the process of computing the loss function is not straightforward. $Z^\pi (s, a)$ and $Y(s', a')$  have disjoint supports since the target distribution supports are modified by $R + \gamma Z^\pi (s', a')$. That is,  the discount factor $\gamma$ shrinks the distribution, while the reward $R$ shifts it. Therefore, the minimization of the KL divergence is not always directly possible, needing a projecting or approximation step to match the target supports onto the current prediction supports, see Figure \ref{fig:categorical}.

    \subsection{TD3}
        % Explain here in a direct/short clear way everything about TD3
        The Twin Delayed Deep Deterministic (TD3) approach introduced by \cite{fujimoto2018addressing} is an actor-critic algorithm built on the DDPG \cite{barth2018distributed} for continuous control problems. TD3 not only improves the results with respect to DDPG but also reduces the sensitivity to the hyperparameters. 
        The key improvement of TD3 is how the overestimation is mitigated. TD3 learns two (scalar) Q-function approximations instead of one, as DDPG does, but it uses the minimum value of the two Q-approximators,  $min (Q_{\theta1}, Q_{\theta_2})$, to calculate the Bellman error loss function. This process leans toward the underestimation of Q values. That is, $TD= (Y - Q_\theta(s, a))$ where $Y = R + \gamma   min_{i=1,2}Q_{\theta_i}(s', a')$.
        Inspired by Double Q-learning \cite{van2016deep}, TD3 also includes target networks (for both actors and critics) to promote stability during policy training. Further, TD3 updates the actor network (i.e., the policy) less frequently than the two critic networks (i.e. Q-function approximators). This allows the critic networks to become more stable and reduce errors before it is used to update the actor network. The target networks are soft-updated and step-delayed with respect to the critics. 
        
        The policy $\pi$ is learned using the deterministic policy gradient by maximizing one of the Q approximators. It is unclear why the authors use just the $Q_{\theta1}$, completely ignoring the other Q approximator. A random noise is also included in the target action. This makes the policy avoid exploiting actions with high Q-value estimates. In the same way, to encourage exploration during the training, random noise is included in the environment's actions following the policy.

%___________________________________________________________________________________-%

\section{CTD4}
    
    We present CTD4 that builds upon TD3 and follows the same training dynamics. However, in CTD4, the critic network structure is modified to output the $\mu$ and $\sigma$ that parameterize the normal distribution of the $Z$ approximation, making it a continuous distributional RL algorithm. See Figure \ref{fig:diagram} for a full graphical representation of the proposed architecture.

\begin{figure}
        \centering
            \includegraphics[width=\linewidth]{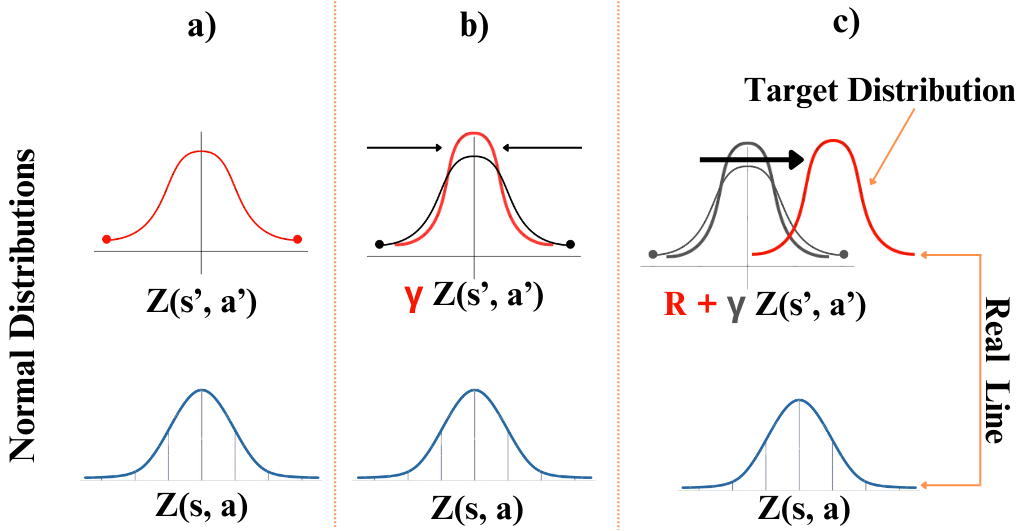}
            \caption{The top part represents the sequence of the Z target distribution being shrunk and shifted by the discount factor $\gamma$ and Reward $R$, respectively. However, since these are continuous distributions, no projection steps or estimations of distance from the nearest support are needed. Therefore, the distance between the target and the current distribution can be calculated directly.  }
            \label{fig:normal}
    \end{figure}

    To improve the Z estimation and, consequently, the policy, an ensemble of $N$ critics are trained to mitigate the overestimation issue. CTD4 distinguishes itself from comparable approaches \cite{choi2019distributional, kuznetsov2020controlling}, which integrate approximations through either averaging or selecting the minimum value among them. In contrast, we employ a distinctive method by fusing the ensemble of critics through a Kalman filter \cite{kalman1960new}. This integration process yields the output parameters $\mu_k$ and $\sigma_k$, that parameterize the normal distribution of the $Z^\pi$ approximation. More specifically, we propose to train the critics $Z_{\theta n}$ for $n \in[1..N]$ of the policy-conditioned return distribution $Z^\pi$ where  each  critic $Z_{\theta n}$ maps the current $(s, a)$ to a continuous distribution parameterize by $\mu_n$ and $\sigma_n$ and the fusion of the critics is calculated by:
    \begin{align}
        k          &= \frac{\sigma_n^2}{(\sigma_n^2 + \sigma_{n+1}^2)} \\
        \sigma_k^2 &=  (1 -k ) \ \sigma_n^2 + k \sigma_{n+1}^2 \\
        \mu_k      &= \mu_n + k (\mu_{n+1} - \mu_n)
    \end{align}
    %
    %
    % Computation of the target distribution
    To minimize the TD error, characterized as the discrepancy between distributions in distributional RL, and facilitate the training of the approximators $Z_{\theta n}$ for $n \in [1..N]$, it is imperative to formulate the Bellman target function as a normal distribution. Specifically, we express the target distribution as $Z_{\text{target}} = R + \gamma \ Z(\mu_k, \sigma_k)$. Since we are assuming a Gaussian normal distribution  $X \sim \mathcal{N}(\mu,\sigma)$, the linear Gaussian transformation rule $Y=aX+b$ can be applied; therefore the target distribution can be defined as:
    \begin{align}
        Z_{target}   &=  R + \gamma \ Z(\mu_k, \sigma_k)\\
        \mu_{target} &\rightarrow \gamma \mu_k + R \\
        \sigma_{target} &\rightarrow \gamma \sigma_k \\
        Z_{target} &= \mathcal{N}(\mu_{target}, \sigma_{target})
    \end{align}
    %
    % This is the strong point of the paper
    A crucial aspect to note for CTD4 is the utilization of continuous distributions instead of categorical distributions. Consequently, the support of the distribution extends along a real continuous line. This characteristic enables us to engage directly with the target distributions, eliminating the necessity for any projection, truncation, or transformation steps. See Figure \ref{fig:normal} for a visual representation.
    The objective of the loss function is to minimize the distance metric between the current approximation distribution and the Bellman $Z_{\text{target}}$ distribution. To quantify the dissimilarity between these distributions, we employ the KL divergence. This parallels the conventional RL framework, where the loss function is designed to minimize the TD error. Consequently, our approach directly minimizes the KL distance between each $Z_{\theta n}$ for $n \in [1..N]$ and the Bellman $Z_{\text{target}}$ distribution as:
     \begin{align}
        TD_{error} &= D_{KL} (Z_{current}, Z_{target}) \\
        TD_{error} &= D_{KL} (Z(\mu_i, \sigma_i), (R + \gamma \ Z(\mu_k, \sigma_k)) \\
        TD_{error} &= D_{KL} (\mathcal{N}(\mu_{i}, \sigma_{i}), \mathcal{N}(\mu_{target}, \sigma_{target}))
    \end{align}
    %--------
    
    It is essential to highlight that despite the introduction of the ensemble-based architecture, a deterministic policy actor $\pi(s)$ remains used. In other words, the actor network produces a deterministic scalar value vector as its output. Specifically, the actor network, characterized by parameters $\phi$, takes the current state $s$ as input and generates the corresponding action to maximise the long-term reward. The Deterministic Policy Gradient \cite{silver2014deterministic} is employed for updating the actor parameters at each training step to maximize the expected discounted reward. Therefore, when considering a sample mini-batch of $M$ transitions for the fusion of the ensemble:
    \begin{equation}
        \nabla_{\phi}J(\phi) \approx \frac{1}{M} \sum_{i=1}^{M} \nabla_a\ \mu_k \nabla_\phi \pi_\phi (s_t) 
    \end{equation}
    
    To encourage exploration, TD3 disturbs the actions selected by the policy through the incorporation of fixed stochastic random noise at each training step $a \backsim \pi_\phi (s) + \varepsilon$.
    Incorporating this idea may prove essential during the preliminary phases, we assert that its continued application may pose a detriment to policy optimization in the final stages of training, particularly in tasks demanding heightened precision. Consequently, we propose a decrease in the magnitude of random noise exploration added to the action. This entails a progressive attenuation of the noise level, denoted as $\varepsilon \backsim \mathcal{N}(0, \sigma)$, at each time step. This strategic adjustment aims to encourage a more stable learning policy, minimizing disturbances.
    
    The same concept as TD3 is followed for the target networks, wherein we periodically update the target parameters $\theta$ and $\phi$ employing the most recent critic and actor parameter values, respectively. In alignment with the idea of decreasing random noise, we extend this concept to the target actions too. 
    %--------
    \subsection{Optimal Fusion Strategy: Why a Kalman Fusion?}
    
        Utilizing an ensemble of approximations has been a recurring strategy in deterministic and categorical RL methods to address overestimation issues inherent in $Q$ approximations. Common practices involve selecting the minimum value among Q-Values \cite{fujimoto2018addressing, lan2020maxmin} or averaging all the approximations \cite{kuznetsov2020controlling, chen2021randomized} as methods to process the output of the ensemble critics. A clear example of this is REDQ which averages the predictions from the ensemble. However, REDQ is limited to deterministic setups and cannot handle distributional updates, whether categorical or continuous. Moreover, its high update rate (G=20) significantly escalates computational costs and risks overfitting without diverse training data. Additionally, although employing an ensemble approach, REDQ only employs two critics randomly selected from the ensemble.% potentially underutilizing the ensemble's power.
        We argue that these approaches have limitations. Opting for the minimum value prioritizes a singular value, underutilizing the potential power of the ensemble. On the other hand, taking the average assigns equal significance to all approximations, irrespective of their potential sub-optimally or inaccuracy.
        
        \begin{figure}
        \centering
            \includegraphics[width=\linewidth]{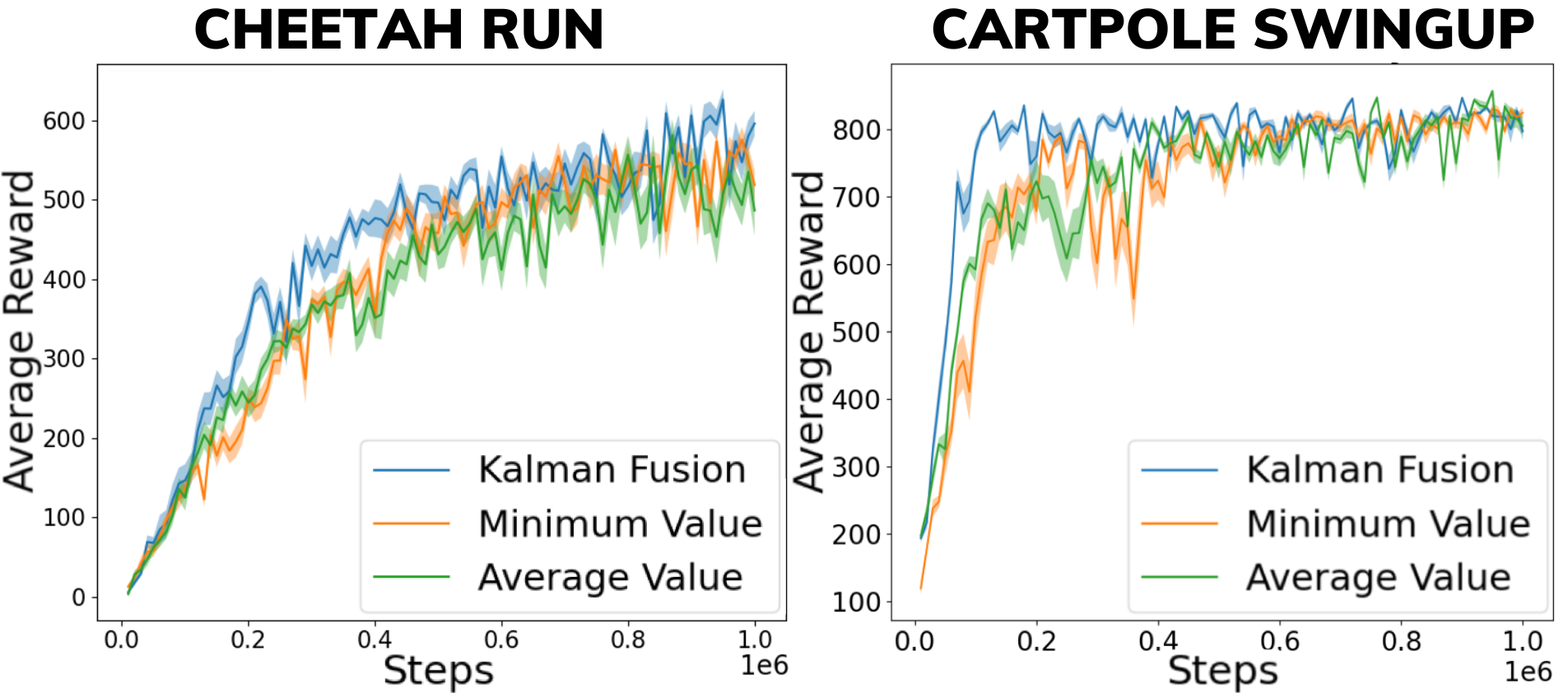}
            \caption[Fusion Method Analysis]{Fusion Method Analysis: Kalman Fusion (Proposed) versus Minimum and Average Value Methods in two environments.}
            \label{fig:Fusion_Analysis}
        \end{figure}

        Notably, in our proposed framework, the output of each critic is a normal distribution parameterized by $\mu$ and $\sigma$. This perspective allows us to interpret each approximation as a sensor reading, enabling the application of sensor fusion methods. To enhance Q-value estimation and fully leverage the collective strength of the ensemble, we employ a Kalman filter to fuse multiple distribution approximations. 
        The Kalman fusion identifies the most probable approximation within the ensemble by calculating the joint probability of the distributions. This strategic fusion optimises Q-value estimates and endows the RL agent with resilience to noise, errors, or inaccuracies inherent in the individual approximations. It is crucial to emphasize that we do not keep track of all the previous approximations; we only fuse them in each training step. 

        Under identical conditions, only the ensemble fusion method is altered among Kalman fusion, minimum value fusion, and average fusion. The agent is then trained for a fixed number of steps.  The results for two complex tasks from the DMCS (Cheetah run and Cartpole Swingup) are displayed in Figure \ref{fig:Fusion_Analysis}. Each fusion method for each task was trained using five different seeds. The results indicate that the proposed method, employing Kalman fusion, exhibits greater stability and achieves higher rewards with fewer samples compared to other fusion methods.

    \subsection {Ensemble Size Optimization: Determining the Optimal Number of Critics}

        \begin{figure}
        \centering
            \includegraphics[width=\linewidth]{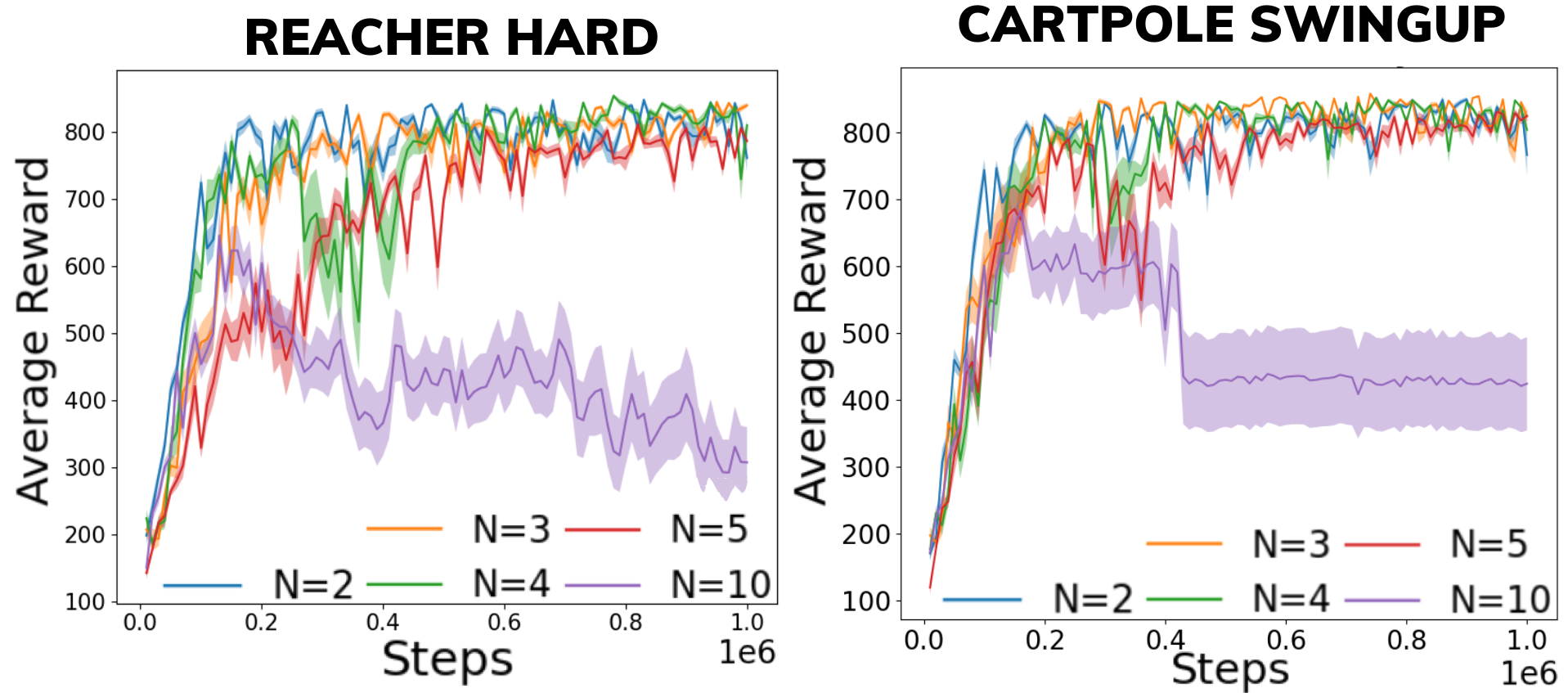}
            \caption[Ensemble Size Analysis]{Ensemble Size Analysis Varying Number of Critics Under Identical Conditions.}
            \label{fig:Ensemble_analysis}
        \end{figure}

        Selecting an appropriate number of approximations is crucial for ensuring stability throughout the training process. Previous related work, such as TD3, does not explain the number of approximators chosen in the ensemble and their effect on the training.
        
        We conducted an analysis to comprehend the influence of ensemble size. Increasing the number of critics can significantly mitigate overestimation, leading to a potential enhancement in overall performance. In other words, a higher number of approximations corresponds to a more pronounced alleviation of overestimation \cite{lan2020maxmin, kuznetsov2020controlling}. Nevertheless, if the number of critics becomes excessively high, there is an elevated risk of instability, potentially leading to underestimation, discouraging exploration, and, consequently, yielding a suboptimal policy \cite{lan2020maxmin}. Additionally, the computational cost experiences a substantial increase with the growing number of critics in the ensemble.
        To determine the correct ensemble size, an agent is trained under the same conditions, using a Kalman fusion for the approximation but varying only the number of approximators. The analysis results, conducted across five different seeds for two complex unrelated tasks (Reacher Hard and Cartpole Swingup), are illustrated in Figure \ref{fig:Ensemble_analysis}. Suboptimal performance may occur if the number of critics is too high (for instance, \(N=10\)). Based on these results, an ensemble  \(N=3\) was selected.

\section{Experiments}

    Our experimental testing aims to evaluate the sample efficiency and ease of training of Continuous Distributional approximators for continuous action spaces. We conduct comparative analyses against TD3 and REDQ, using the official source-code implementations provided by the authors. REDQ is selected for its similarity to using an ensemble method, aligning with the approach taken in this article. TD3 is chosen for its well-established dominance across a range of algorithms designed for continuous action spaces.  
    
    We purposefully avoid applying or comparing our method with previous approaches, such as C51 or IQN, which are tailored for discrete action spaces. Our research squarely focuses on addressing the challenges posed by complex tasks characterized by continuous action spaces—a domain notorious for its heightened difficulty level. Additionally, we do not compare our proposal against TD4 and DSAC, as they rely on parallelized learning. Instead, our focus is on improving learning efficiency in a single-agent setup, which is more practical and resource-efficient.

    To conduct experiments with our proposed method, we select the challenging continuous control tasks from the DeepMind Control Suite (DMCS) \cite{tassa2018deepmind}.
    We choose ten environments, each presenting distinct levels of complexity and task requirements. The default reward signal provided for each environment ensures consistency and comparability across evaluations. The experiments are conducted over a training duration of $1\times10^6$ steps for each environment, employing identical hyperparameters for five independently seeded runs. We regularly evaluate the agent's performance in each environment during the training; after every $1 \times 10^4$ training steps, we compute the average reward over $10$ consecutive episodes. It is important to note that no gradient updates are executed during the evaluation phase. Notably, all model parameters are optimized using the Adam optimizer, with a single gradient update executed per environment step. The training process was completed on a PC with an i9 CPU, 128GB of RAM, and a GeForce RTX4090 NVIDIA GPU with Ubuntu 20.04. 

    To facilitate the replication of our work, we provide all raw data plots corresponding to each independent seed, a comprehensive list of hyperparameters, a training loop, the entire source code, and task demonstration videos. This supplementary material is available on the paper's website at: \url{https://sites.google.com/view/ctd4}.

\section{Results and Discussion}

\begin{figure*}
    \centering
     \begin{tabular}{@{}ccc@{}}
     
    \includegraphics[width=.315\textwidth]{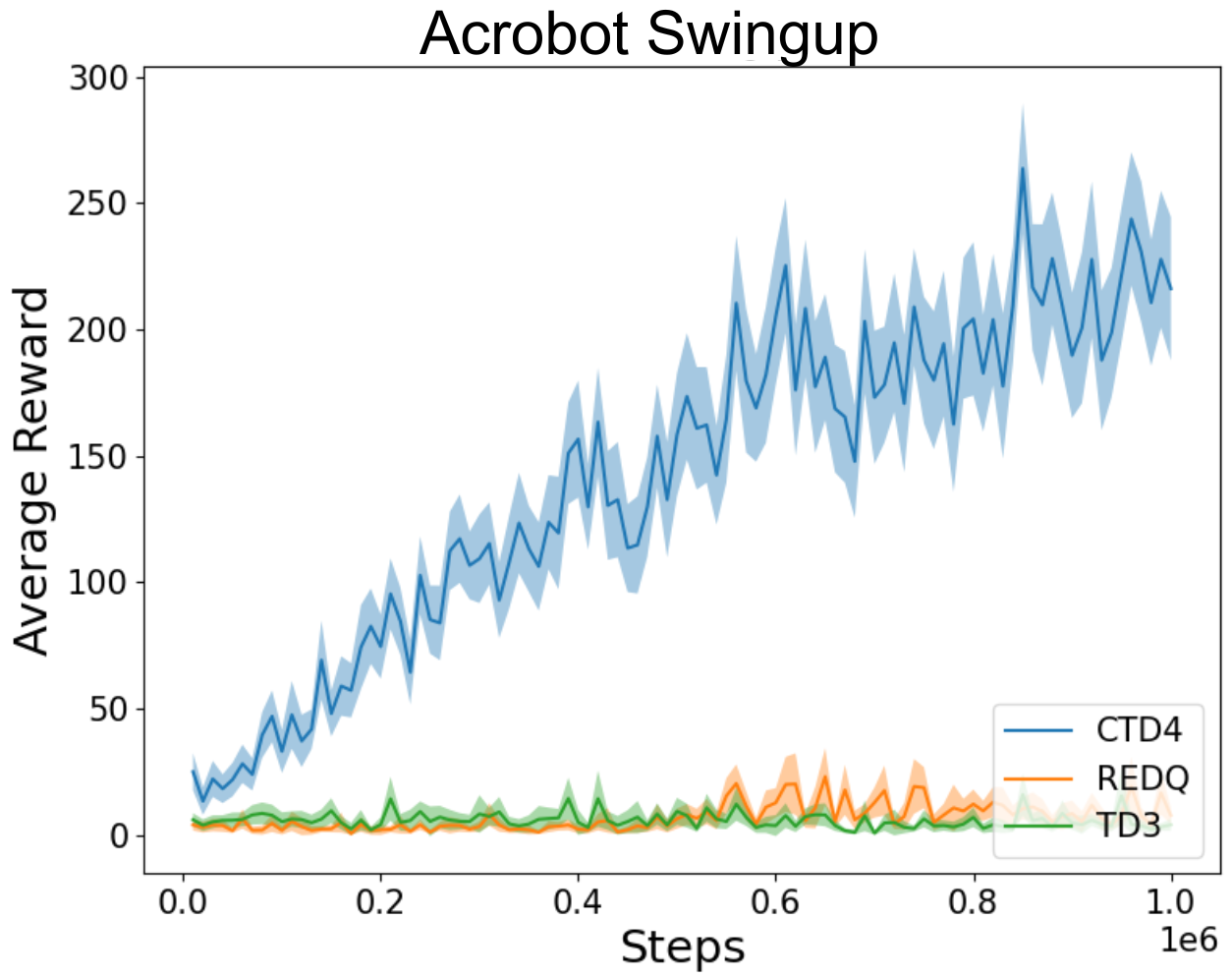}&
    \includegraphics[width=.315\textwidth]{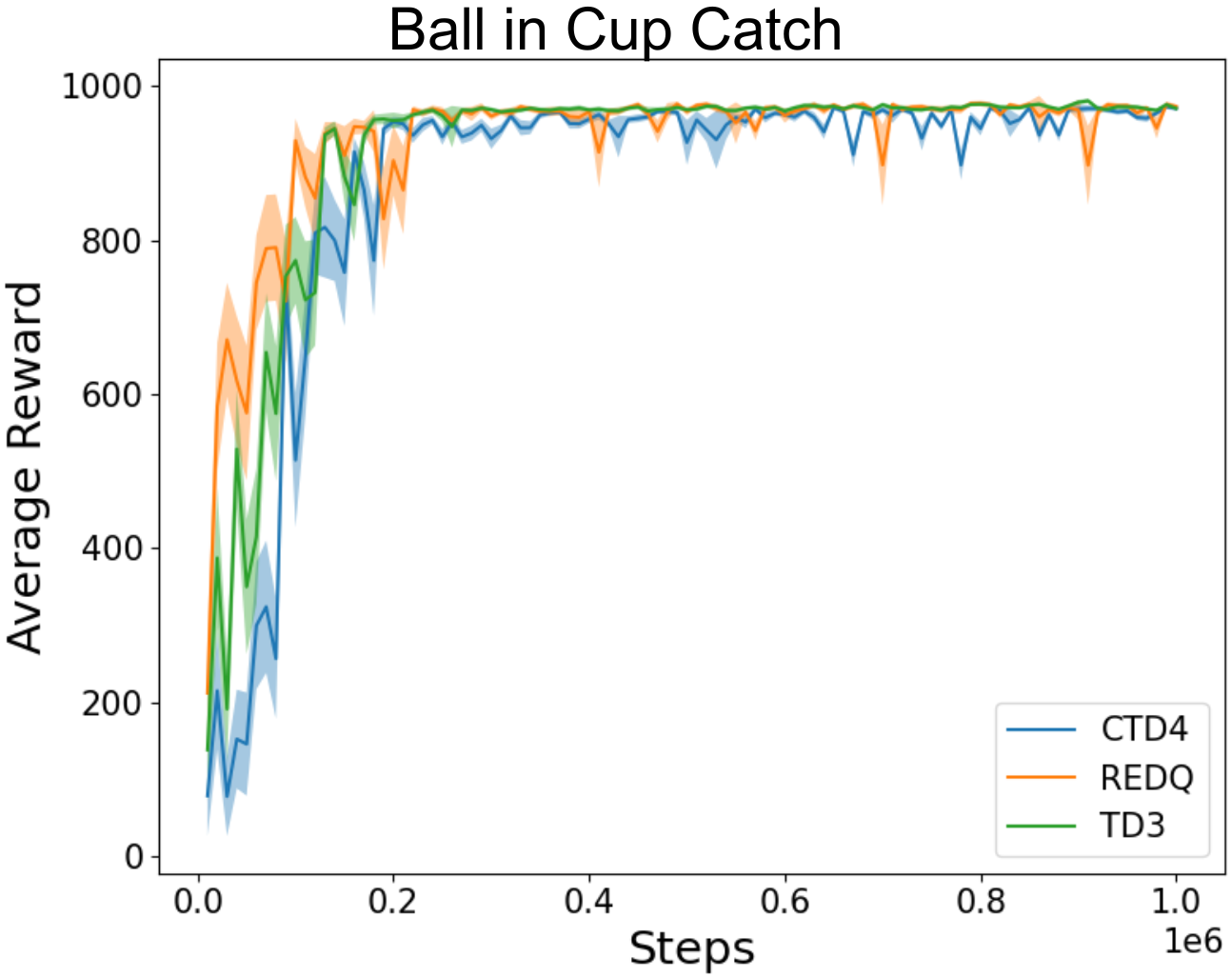} & 
    \includegraphics[width=.315\textwidth]{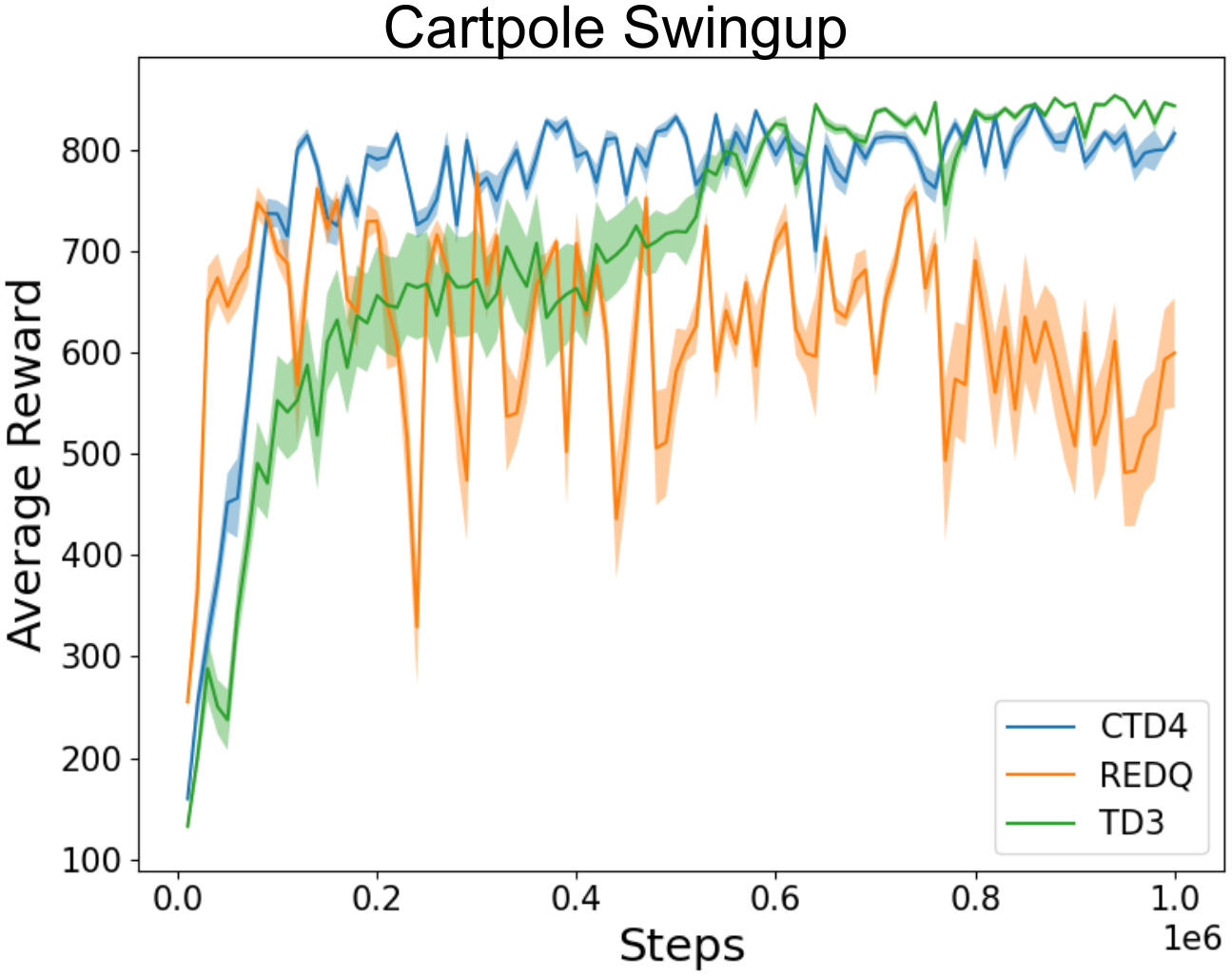} \\
    \includegraphics[width=.315\textwidth]{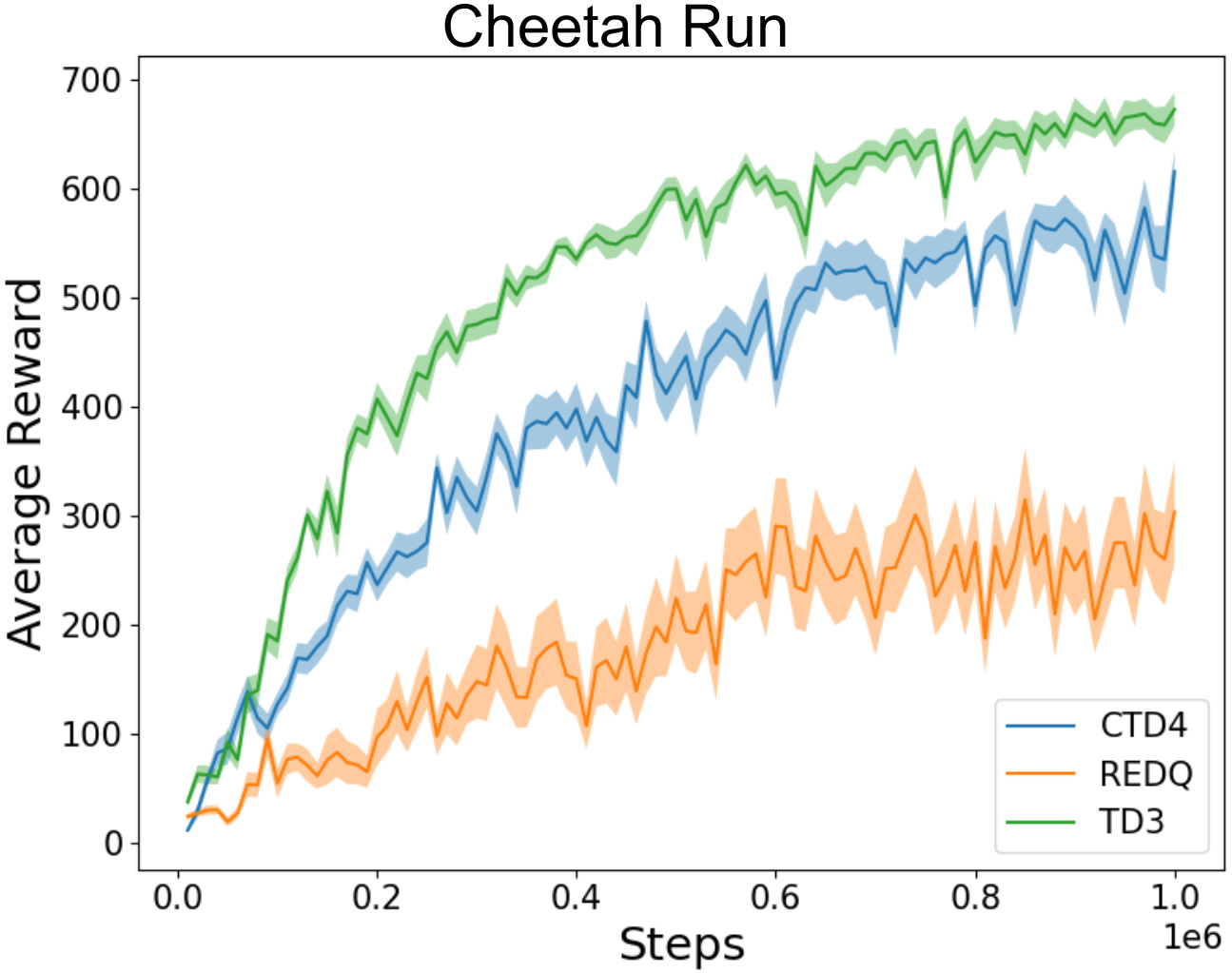} &
    \includegraphics[width=.315\textwidth]{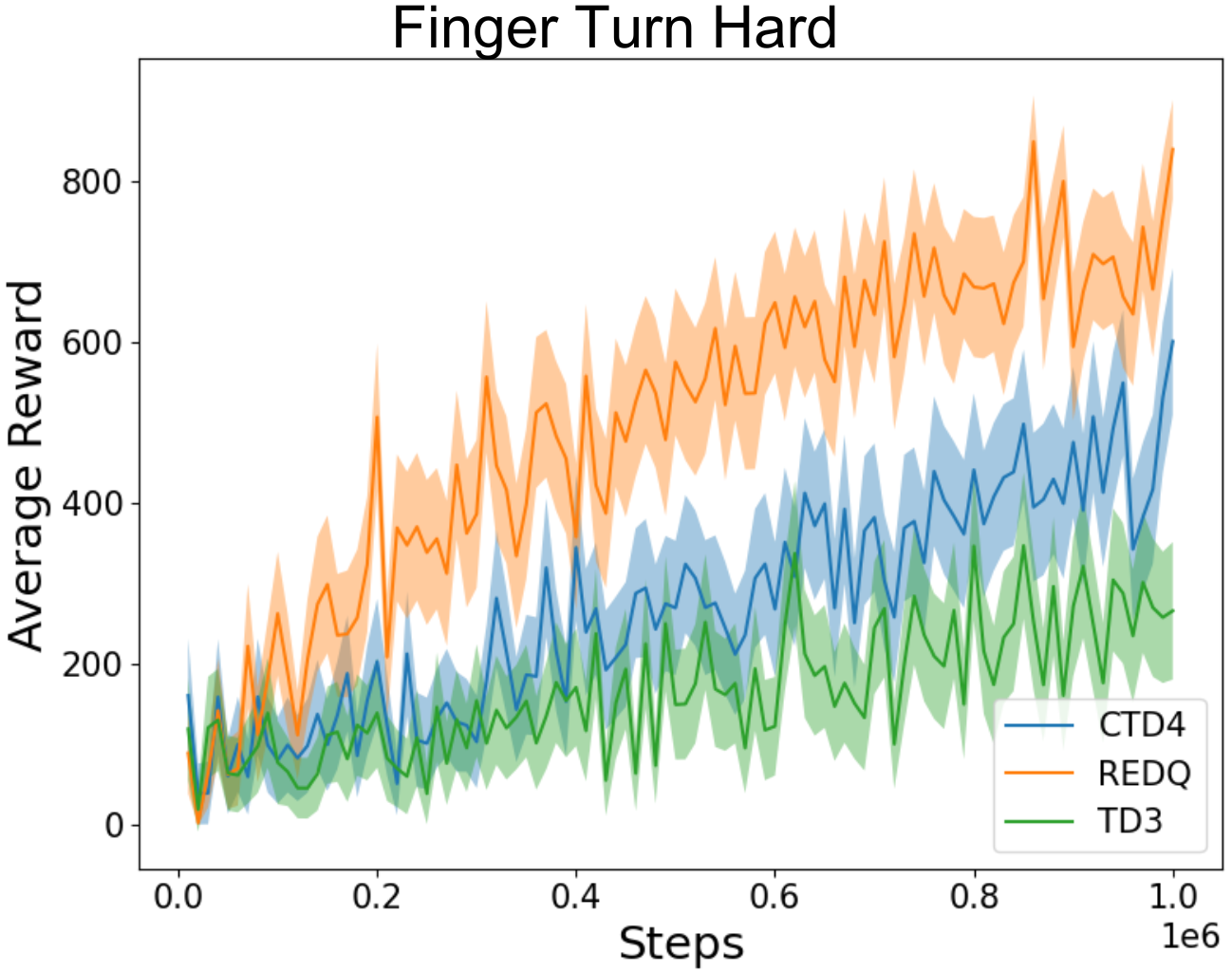} &
    \includegraphics[width=.315\textwidth]{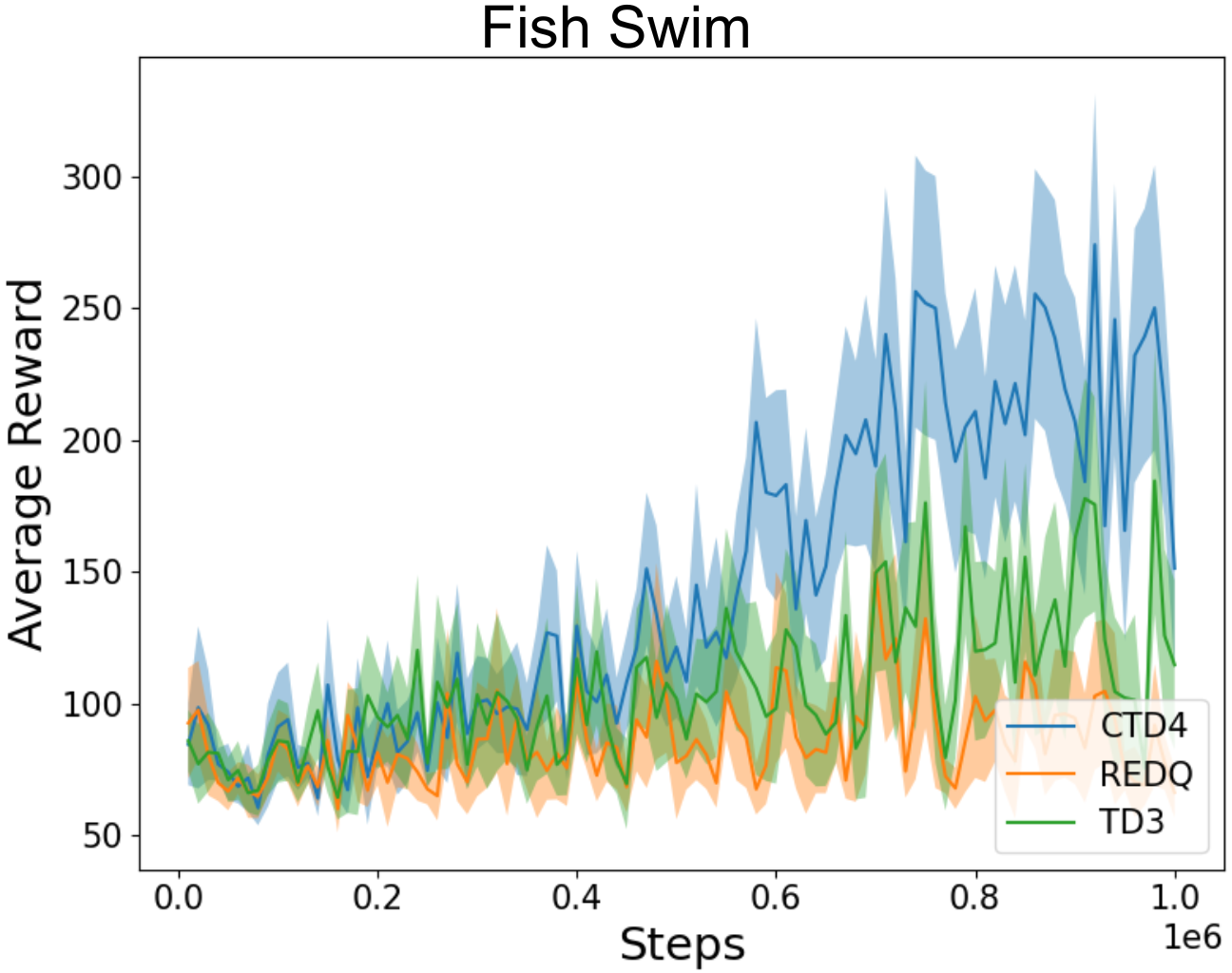} \\
    \includegraphics[width=.315\textwidth]{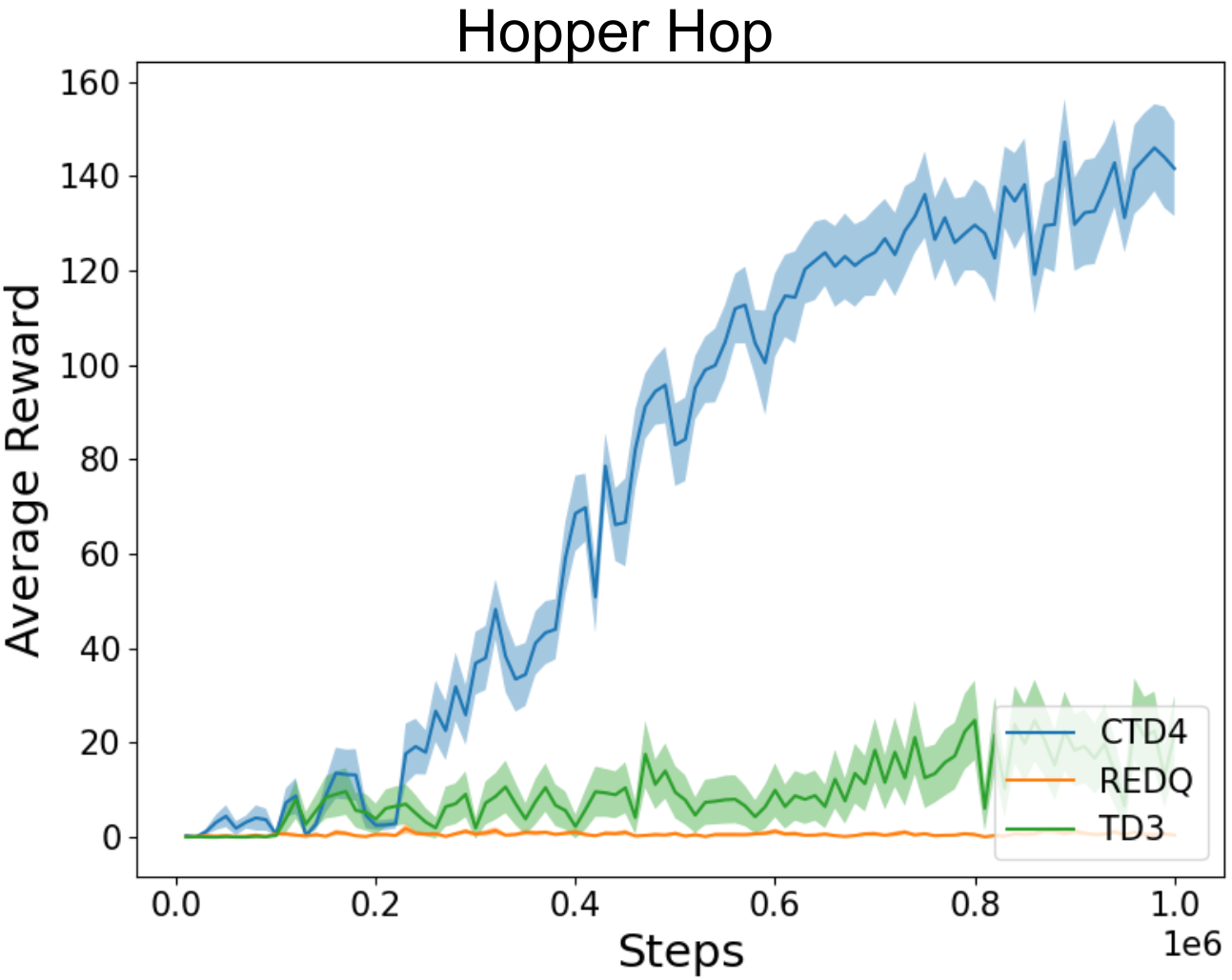} &
    \includegraphics[width=.315\textwidth]{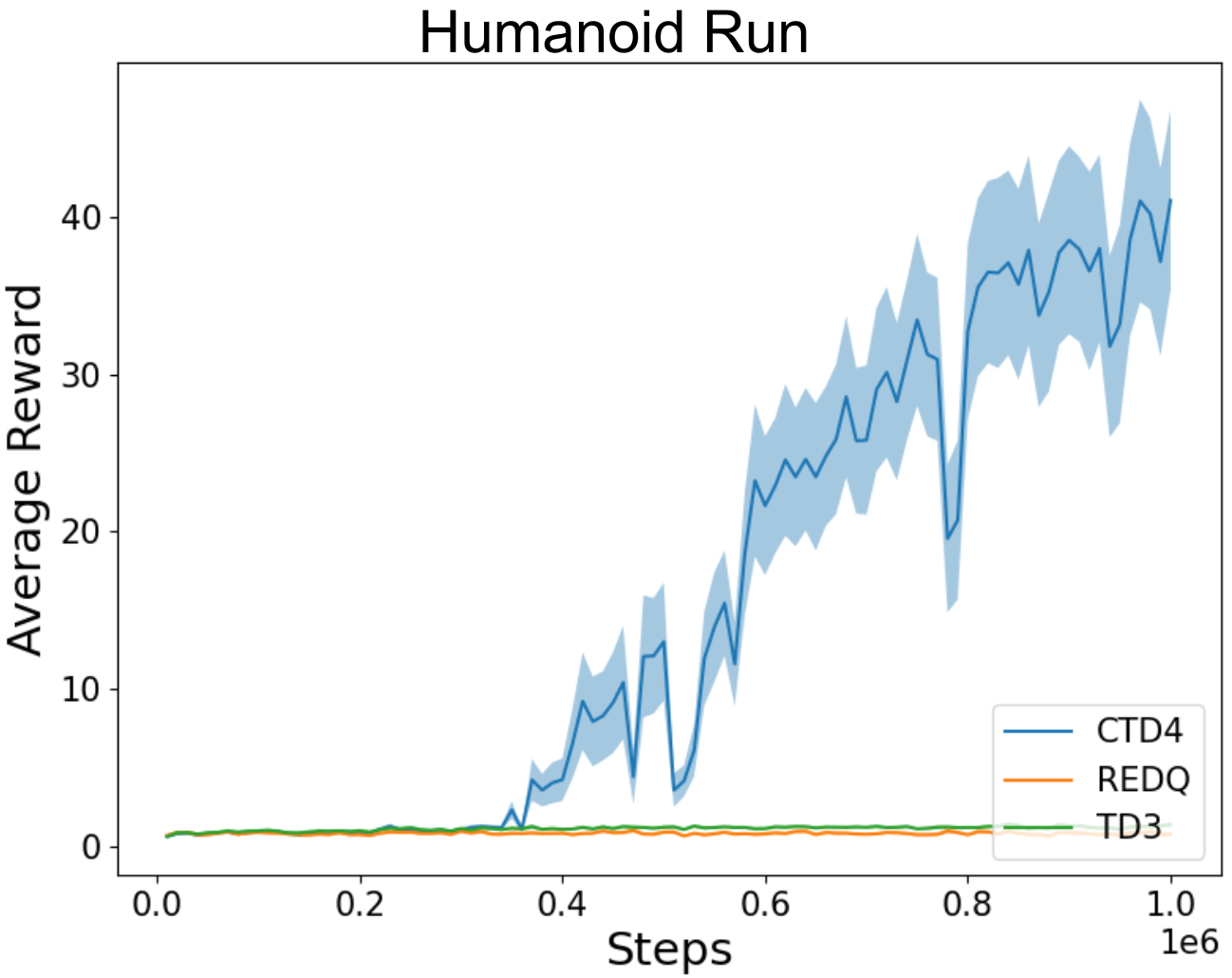} &
    \includegraphics[width=.315\textwidth]{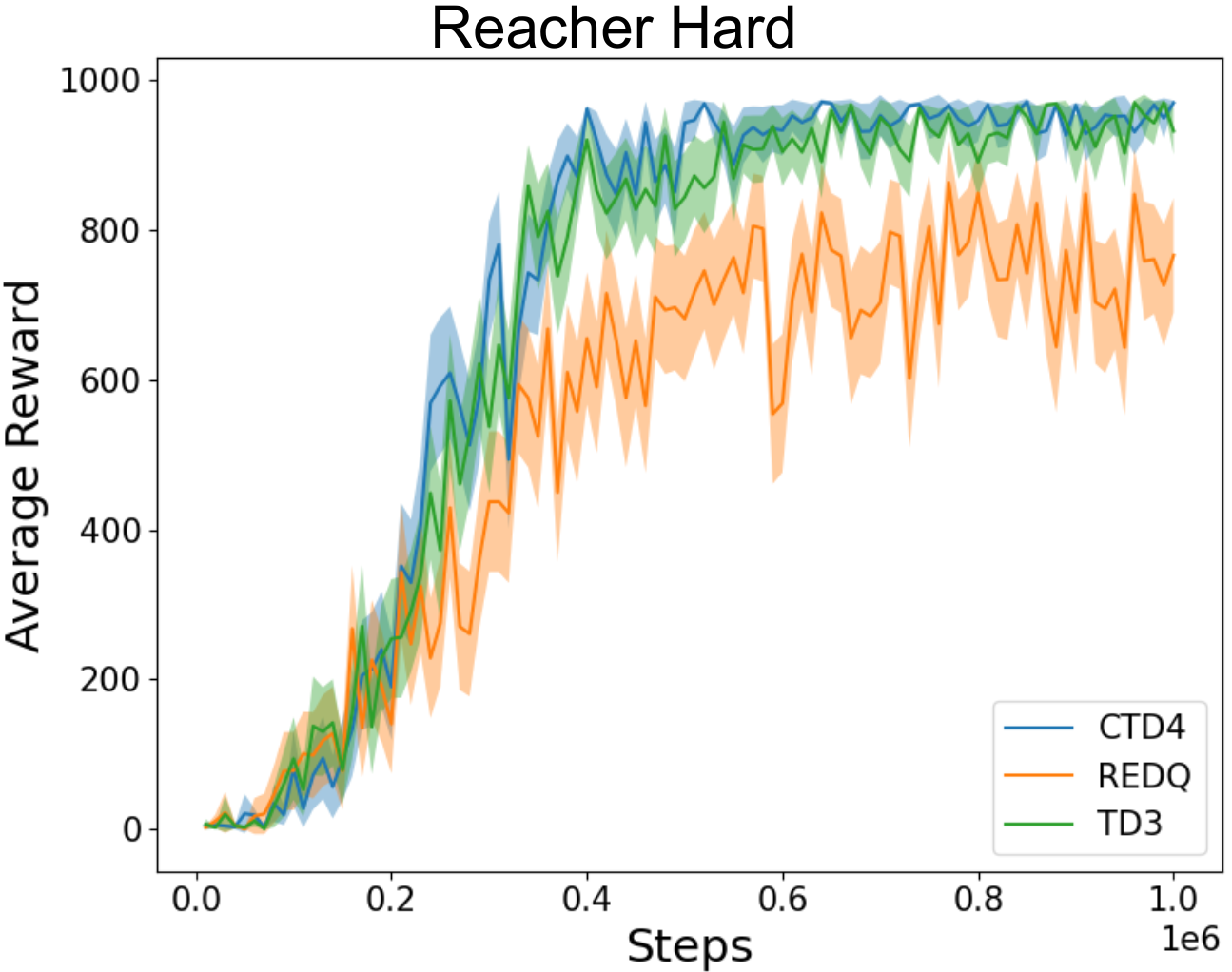} \\
    \multicolumn{3}{c}{\includegraphics[width=.315\textwidth]{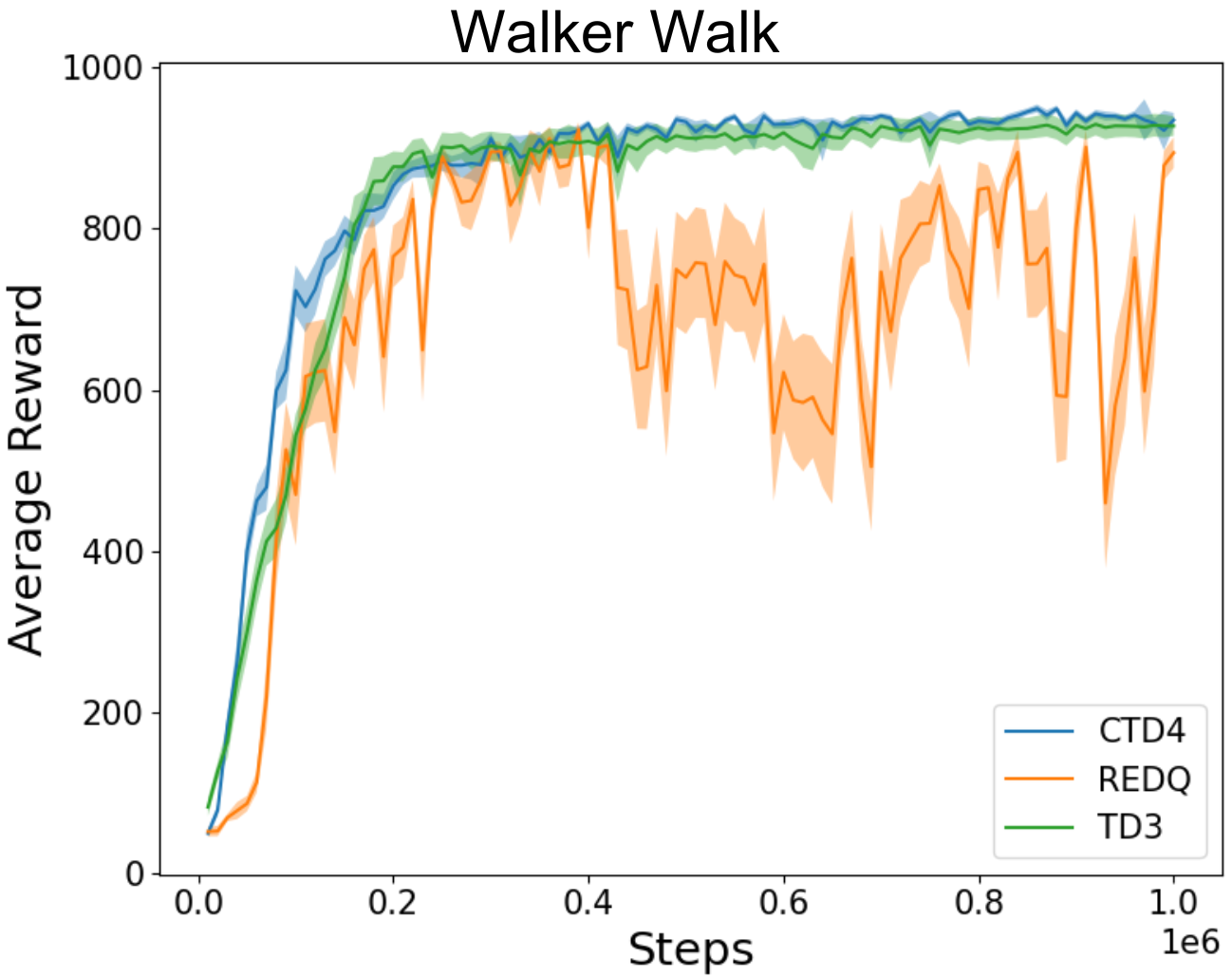}}
  \end{tabular}
  \caption{Learning curves for CTD4, REDQ and TD3 on ten DeepMind Control Suite continuous control tasks. In each plot, the solid line represents the average, while the shaded region represents the standard deviation of the average evaluation over five independent seeds.}
  \label{fig:result_plots}
\end{figure*}

    Figure \ref{fig:result_plots} shows the average evaluation reward across multiple seeds, comparing the performance of CTD4, TD3, and REDQ under identical conditions.

    Using a continuous distribution approach parametrized by the mean and standard deviation, coupled with the fusion of multiple critics through a Kalman method, is effective in achieving comparable or enhanced performance across diverse tasks. The superiority of our proposed methodology becomes evident, particularly in complex scenarios where TD3 and REDQ struggled to solve tasks or exhibit patterns of task completion.
    
    It is crucial to note that we deliberately selected exceptionally challenging tasks to assess the performance of our proposal. Specific tasks, such as \textit{Reacher Hard}, \textit{Finger Spin Hard}, or \textit{Ball-in-Cup}, pose increased difficulty due to their sparse reward structures, adding complexity to task-solving. Conversely, tasks like \textit{Walker Walk} and \textit{Humanoid Run} demand intricate coordination among numerous joints, making them inherently complex. The \textit{Acrobot Swingup} task necessitates balance and precise control, while the \textit{Fish Swim} task involves navigating a 3D environment, introducing an additional layer of complexity to the learning process. The results affirm that the stochastic nature of CTD4 can discover superior solutions with higher rewards. This becomes particularly evident in environments such as \textit{Hopper Hop} or \textit{Acrobot Swingup}, where the performance curves demonstrate that our proposal can outperform TD3 and REDQ without the need for complex hyperparameter tuning or intricate projection steps as required by other distributed RL methods.

\section*{Conclusions}

    CDRL has demonstrated efficacy and enhanced the performance of traditional RL. Nonetheless, its practical implementation is complex, marked by numerous computations and tuning steps. A noteworthy limitation in the existing literature is the inability of several CDRL approaches to address complex tasks within continuous action spaces effectively.  In this paper, we set out to investigate the use of continuous distributions as a mechanism for improving and simplifying the use of distribution in RL. Our proposed solution involves leveraging continuous distributions, specifically a normal distribution parameterized by mean and standard deviation. This approach addresses and mitigates the majority of the problems associated with CDRL, facilitating a more straightforward implementation process. Importantly, our method bypasses the need for custom loss functions, as we minimize the KL divergence between two distributions, a reasonably manageable loss metric to minimize.

    Our study additionally addresses the prevalent issue of overestimation bias in actor-critic RL methods. To tackle this challenge, we propose utilizing an ensemble of multiple approximations (critics) integrated through a Kalman approach. Our findings reveal that the Kalman fusion effectively mitigates overestimation, surpassing the limitations associated with conventional fusion methods, including average fusion or minimum value selection. Likewise, we introduce a noise decay approach combined with the actions generated by the policy. Our results illustrate that the random noise decay approach effectively promotes exploration in the early stages, and the gradual reduction ensures that it does not affect learning in the concluding stages of the training. This idea can greatly improve performance, especially in environments that need high precision. Finally, our modifications and contributions are straightforward to integrate and implement across other actor-critic algorithms transitioning from deterministic to distributional RL.

    Future work will focus on deploying this implementation in real-world scenarios that demand a stochastic nature and involve complex situations, such as robot bipedals or dexterous hand manipulations. Further analysis is also crucial, exploring diverse continuous probability distributions and their impact on policy, opening the door for future research in the realm of continuous distributions RL.
    
    %---------------

\section*{Acknowledgements}
This research was partially supported by the New Zealand Ministry for Business, Innovation and Employment (MBIE) on contract UOAX1810.

\bibliography{bibliography}

\end{document}